\pdfoutput=1

\documentclass[final,5p,times,twocolumn]{elsarticle}

\usepackage{amssymb}
\usepackage{lineno,hyperref}
\usepackage{multirow}
\usepackage{amsmath}
\usepackage{amsthm}
\usepackage[small]{caption}
\usepackage{graphicx}
\usepackage{color}

\journal{arxiv}

\begin{document}

\begin{frontmatter}



\title{Attentive Multi-View Deep Subspace Clustering Net\tnoteref{label0}}
\tnotetext[label0]{This work was supported by}

\author[label1]{Run-kun Lu}
\ead{zsylrk@gmail.com}
\author[label1]{Jian-wei Liu}
\ead{liujw@cup.edu.cn}
\author[label1]{Xin Zuo}
\ead{zuox@cup.edu.cn}

\address[label1]{Department of Automation, College of Information Science and Engineering, China University of Petroleum, Beijing, 260 Mailbox, Changping District, Beijing 102249, China}

\begin{abstract}
In this paper, we propose a novel Attentive Multi-View Deep Subspace Nets (AMVDSN), which deeply explores underlying consistent and view-specific information from multiple views and fuse them by considering each view's dynamic contribution obtained by attention mechanism. Unlike most multi-view subspace learning methods that they directly reconstruct data points on raw data or only consider consistency or complementarity when learning representation in deep or shallow space, our proposed method seeks to find a joint latent representation that explicitly considers both consensus and view-specific information among multiple views, and then performs subspace clustering on learned joint latent representation.Besides, different views contribute differently to representation learning, we therefore introduce attention mechanism to derive dynamic weight for each view, which performs much better than previous fusion methods in the field of multi-view subspace clustering. The proposed algorithm is intuitive and can be easily optimized just by using Stochastic Gradient Descent (SGD) because of the neural network framework, which also provides strong non-linear characterization capability compared with traditional subspace clustering approaches. The experimental results on seven real-world data sets have demonstrated the effectiveness of our proposed algorithm against some state-of-the-art subspace learning approaches.

\end{abstract}




\end{frontmatter}


\section{Introduction}
\label{sec 1}
In recent years, multi-view learning has attracted widespread attention from machine learning researchers because it usually considers more complete information than single-view algorithms. Generally, the field is consisting of multi-modal and multi-feature learning, where multi-modal emphasizes that data come from multiple modalities (such as video, audio, and text); multi-feature emphasizes that data obtained by different feature extraction methods. Most existing so called multi-view algorithms are belong to multi-feature learning \cite{DBLP:journals/corr/abs-1908-01978}, and this paper is also under the assumption of multi-feature data. On the other hand, in this field, multi-view self-representation subspace learning is one of the most attractive directions, where most existing multi-view subspace clustering algorithms are actually self-representation based methods \cite{Zhang2015,7298657,8099944,Brbic2018,Luo2018,8740912}. 

Self-representation assumes that each instance in input space can be expressed by linear weighted combination of other instances, and given the instances set $\mathbf{X}\in \mathbb{R}^{M\times N}$, the process can be formulated as: 
\begin{equation}
\begin{array}{l}
\mathop{\min}\limits_{\mathbf{C}}R\left( \mathbf{C} \right)\\
s.t.\;\mathbf{X}=\mathbf{XC},\;\mathbf{C}\in \Omega  ,\\
\end{array}
\label{eq0}
\end{equation}
where $\Omega$ is some matrix set, and $R\left( \cdot \right) $ is the regularization of self-representation matrix which can implement different constraints on optimal problem (usually frobenius, $L_1$, nuclear norms are alternatives). Besides, ${\bf{C}} \in  {\mathbb{R}^{N \times N}}$ is the self-representation matrix, where ${\left[ {\bf{C}} \right]_{i,j}}$ measures the correlation between instance $i$ and $j$, and then the self-representation matrix can be utilized to construct affinity matrix of spectral clustering according to
\begin{equation*}
\frac{\left| \mathbf{C} \right|+\left| \mathbf{C}^{\text{T}} \right|}{2}.
\end{equation*} 
Similar to other multi-view learning algorithms \cite{Liu2015,Lu2020}, multi-view subspace clustering also focus on multiple views' consistency and complementarity, where consistency represents views' consistent subspace which can be achieved by low-rank regularization \cite{8099944} or views alignment \cite{Brbic2018}, and complementarity usually focus on the view-specific subspace which can be achieved by frobenius regularization \cite{Luo2018}.

Subspace clustering works well under the assumption that sample space has relatively separable decision boundaries, when facing highly nonlinear data, only considering the regularization of self-representation matrix can not effectively improve the clustering performance due to the limitation of input representation. Based on this, latent multi-view subspace clustering \cite{8099944} first computed a latent representation of multiple views, and then applied it in subspace clustering; more recently, multi-view low-rank sparse subspace clustering \cite{Brbic2018} has proposed kernel-based variants that first project input into high dimensional kernel space. However, the first approach essentially obtained the latent feature through a linear transformation, and the second one encountered the problem of kernel functions selection. To mitigate these problems, neural network is considered more suitable for its great non-linear and adaptable (without the kernel functions selection) ability, such as Deep Subspace Clustering Networks \cite{DBLP:journals/corr/abs-1709-02508} and Deep Multimodal Subspace Clustering Networks (DMSCN) \cite{8488484}, they added additional self-representation layer in fully convolutional auto-encoders to achieve deep subspace learning. However, they are available only for 3-D tensor (we view gray-scale image as one channel but still 3-D tensor, such as for MNIST the dimension is [28, 28, 1]), but for most multi-view data, each view is always a kind of 1-D hand-craft feature of raw data that the fully convolutional network is not suitable for this case; and DMSCN is a multi-modal model but not considers different modal's contribution in modal fusion process. 

\textcolor{black}{In multi-view learning task, different data uses various methods when constructing multiple views, but there is no universal criterion to measure the quality of view representation. Compared to relying on experience, an end-to-end approach is necessary to measure the degree of view importance through learning. Fortunately, with the great progress made by transformer \cite{NIPS2017_7181} in NLP \cite{Brown2020LanguageMA} and CV \cite{DBLP:journals/corr/abs-2010-11929} tasks, attention mechanism has become an essential module for many of the recent neural architectures, which allows the model to dynamically pay attention to only certain parts of the input.}
Motivated by this, we propose a novel framework called Attentive Multi-View Deep Subspace Net (AMVDSN), which is a fully connected layer based network. Our proposed method takes both multiple views' consistent and view-specific information into consideration and fuses them with dynamic weights into a joint latent representation through attention, and by utilizing this mechanism, we can effectively exploit the contribution of each view. Based on this, just a simple regularization like frobenius norm of self-representation matrix, which can be optimized automatically in deep learning programming framework, will derive much better clustering performance than traditional self-representation subspace clustering algorithms. On the other hand, AMVDSN is a deep model constructed mainly by fully connected layers, which is easy to encounter the problem of model degradation. Therefore, by introducing shortcut connection \cite{DBLP:journals/corr/HeZRS15}, we can effectively alleviate model degradation and improve clustering performance.

In summary, the main contributions of this paper are:

(1) the proposed algorithm is the first approach to apply self-attention mechanism in multi-view subspace clustering, which incorporates multiple views' dynamic contributions to representation learning and derive more reasonable non-linear joint latent representation consists of consistent and view-specific information;

(2) compare with traditional subspace learning methods that have complicated regularization on self-representation matrix and complex optimization process, we just select simple frobenius norm as regularization term that can be optimized automatically using SGD and derive much better clustering performance;

(3) the proposed algorithm is the first multi-view subspace clustering method that introduces shortcut connection to solve the problem of model degradation.

(4) the proposed method has achieved state-of-the-art clustering performance on some real-world data, especially on Prokaryotic, which improved by more than 10 percentage points on multiple metrics.

Our paper is organized as follows. In section \ref{sec 2}, we introduce some related works on multi-view learning, subspace clustering, and multi-view subspace clustering; In section \ref{sec 3}, we give a detailed description on our proposed algorithm; In section \ref{sec 4}, we demonstrate the effectiveness of our proposed method through some experiments on real world-data sets; In section \ref{sec 5}, we will conclude this paper. 

\section{Related Work}
\label{sec 2}
\subsection{Multi-View Learning}
Recently, multi-view learning has become an important research direction of machine learning, and the field usually focus on utilizing more complete information from multiple views compared with single view algorithms. Canonical correlation analysis (CCA) \cite{10.1093/biomet/28.3-4.321} and co-training \cite{Blum:1998:CLU:279943.279962} are always viewed as the early work of multi-view learning, and scholars have used them as a basis to develop many variants in this field. According to Sun's book \cite{Sun2019} (the first book on multi-view learning), the research topics in this area mainly include: multi-view supervised \cite{10.1007/978-3-319-57529-2_51,8008811} and semi-supervised \cite{QI201246, CHAO2019296} learning, multi-view subspace learning \cite{7072521,7123622}, multi-view clustering \cite{Yang2018,pmlr-v97-peng19a,Huang2019}, multi-view active learning \cite{DBLP:journals/jair/MusleaMK06,DBLP:journals/pr/ZhangS10}, multi-view transfer \cite{doi:10.1002/sam.11226} and multi-task learning \cite{7555376,Lu2020}, multi-view deep learning \cite{DBLP:conf/icml/AndrewABL13}, and view construction \cite{10.1007/978-3-642-21105-8_69}. Besides, the field can be widely used in plenty of applications, such as computer vision \cite{Wang2019}, social network \cite{8119126}, recommendation Systems \cite{elkahky2015a}, and medical research \cite{Zhang2018}.

\subsection{Self-representation Subspace Clustering}
\label{sec 2.2}
In the past decade, subspace clustering has been widely used in many real-world applications \cite{DBLP:conf/cvpr/RaoTVM08,DBLP:conf/cvpr/FengLXY14,DBLP:conf/iccv/JiSL15}, and particularly, self-representation based methodologies have achieved promising clustering results. Such methods belong to spectral clustering, which means they all need to learn a affinity matrix to measure the similarities among different instances, and one of the most effective approaches is self-representation clustering. According to equation (\ref{eq0}), different regularization terms correspond to various subspace clustering algorithms, such as least squares regression (LSR) \cite{10.1007/978-3-642-33786-4_26}, sparse subspace clustering (SSC) \cite{6482137}, and low-rank representation (LRR) \cite{Vidal2014} use frobenius, $L_1$, and nuclear norms respectively. Further more, multi-subspace representation (MSR) \cite{10.1007/978-3-642-23783-6_26} combines $L_1$, and nuclear norms together to utilize the advantages of SSC and LRR. The works discussed above all based on the assumption that self-representation matrix $C$ follows block diagonal property, but actually most existing subspace clustering algorithms have indirect structure priors, and based on this Lu et al. has proposed block diagonal representation (BDR) to deal with this problem \cite{8259470}. Although the above research works have achieved promising performance, they still suffered from the problem that each instance should be linear reconstructed, which is at odds with most complex real-world applications involve nonlinear relationships. Based on this, some researchers have proposed kernel-based approaches \cite{Patel_2013_ICCV,7025576,Yin_2016_CVPR,7283631}, but they still faced the dilemma of kernel selection. Therefore, neural network has naturally become the better solution for its strong non-linear ability without kernel selection problem, some auto-encoder based subspace clustering methods have been proposed and achieved impressive performance \cite{9000790,8387808}.

\begin{figure*} [htb]
	\begin{center}
		\includegraphics[width=0.9\textwidth]{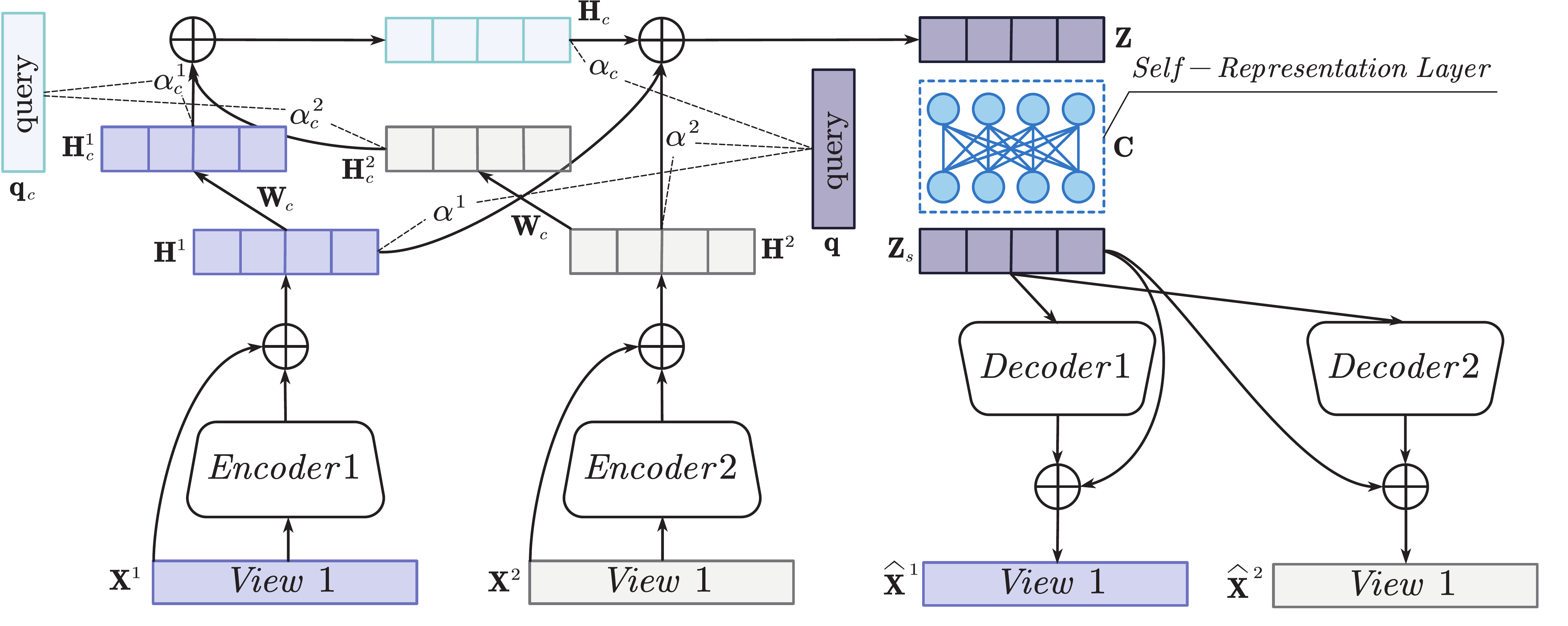}
		\caption{AMVDSN's Framework: Take data with two views as an example. ${\bf{X}}^1$ and ${\bf{X}}^2$ (view 1 and 2) are embedded into the same dimension through Encoder 1 and Encoder 2 with structure of shortcut connection, and the corresponding latent representations are ${\bf{H}}^1$ and ${\bf{H}}^2$ respectively. Further, consistent features of each view ${\bf{H}_c}^1$ and ${\bf{H}_c}^2$ are calculated through share weight ${\bf{W}}_c$, and they can be utilized to compute joint consistent feature ${\bf{H}}_c$ through attention as: ${\bf{H}}_c = \alpha_c^1{\bf{H}_c}^1 +  \alpha_c^2{\bf{H}_c}^2$. Then a global latent representation can also be calculated through attention as: ${\bf{Z}} = \alpha^1{\bf{H}}^1 +  \alpha^2{\bf{H}}^2 + \alpha_c{\bf{H}}_c$. ${\bf{Z}_s}$ is the reconstruction of ${\bf{Z}}$ by self-representation layer, and it is used to reconstruct each original view through Decoder 1 and Decoder 2, which means ${{{\bf{\hat X}}}^1}$ and ${{{\bf{\hat X}}}^2}$ are the reconstructions of ${\bf{X}}^1$ and ${\bf{X}}^2$.} \label{figure_1}
	\end{center}
\end{figure*}

\subsection{Multi-View Subspace Clustering}
Subspace clustering has become one of the most attractive research directions of multi-view learning in recent years, and most existing algorithms are self-representation based subspace clustering. Similar to single view subspace clustering algorithms, the early works are usually matrix factorization based methods but have different constraints on self-representation matrix. However, as explained in section \ref{sec 1}, the regularization terms usually correspond to the underlying physical meanings \cite{Brbic2018,Luo2018} in multi-view learning, such as low-rank regularization focus on the consistency between views \cite{8099944}. Comparing with single view approaches, multi-view subspace clustering needs to consider multiple views information and the views fusion methods usually follow two ways: (1) obtain self-representation matrix of all views and fusion them together as one affinity matrix of spectral clustering \cite{Zhang2015,7298657,Brbic2018}; (2) first compute a latent representation of multiple views and then obtain its self-representation matrix \cite{Luo2018,8099944,Zhang2019}. Unfortunately, these algorithms also face the limitation of linear views reconstruction process, kernel based methods cannot deal with the issue well and neural network based method must be the better solution as discussed in subsection \ref{sec 2.2}.

\section{Attention-based Multi-View Deep Subspace Network}
\label{sec 3}
\subsection{Notations}
For a clearer and more accurate description, we denote that bold uppercase and lowercase characters stand for matrices and \textcolor{black}{column vectors} respectively, and other not bold characters represent scalars. \textcolor{black}{Let $\left[\bf{a};\bf{b}\right]$ denotes the vertical concatenation between vectors $\bf{a}$ and $\bf{b}$.} And the network structure described in the remaining subsections is shown in Figure \ref{figure_1}.

\subsection{Framework Description}
\textbf{Encoder embedding layer:} we define ${{\bf{X}}^v} \in {\mathbb{R}^{{M^v} \times N}}$, $v \in \left\{ {1,2, \cdots ,V} \right\}$ is the collection of instances for the $v$-th view, where $M^v$ and \textit{N} are instance dimension and number of $v$-th view respectively. Due to dimensional differences among views, we need first perform embedding operation to ensure that different views characterize as the same dimension, and the encoder of auto-encoder can accomplish this process:
\begin{equation}
{\bf{h}}_n^v = f\left( {{{\bf{W}}_{\left( {L,v} \right)}}f\left( {{{\bf{W}}_{\left( {L - 1,v} \right)}} \ldots f\left( {{{\bf{W}}_{\left( {1,v} \right)}}\left[ {{\bf{x}}_n^v;1} \right]} \right)} \right)} \right),
\label{eq1}
\end{equation}
where ${\bf{x}}_n^v \in {\mathbb{R}^{{M^v}}}$ is \textit{n}-th instance of ${\bf{X}}^v$, i.e. ${{\bf{X}}^v} = \left\{ {{\bf{x}}_n^v} \right\}_{n = 1}^N$; ${{\bf{W}}_{\left( {1,v} \right)}} \in {\mathbb{R}^{{M^h} \times \left( {{M^V} + 1} \right)}}$ and ${{\bf{W}}_{\left( {l,v} \right)}} \in {\mathbb{R}^{{M^h} \times \left( {{M^h} + 1} \right)}}$ are the combination of fully connected layer's weights and biases, and $l \in \left\{ {2, \cdots ,L} \right\}$ stands for the \textit{l}-th layer of encoder; in addition, $f\left(  \cdot  \right)$ is the activation function, and in this paper we use Relu \cite{pmlr-v15-glorot11a}. Specially, ${\bf{h}}_n^v \in {\mathbb{R}^{{M^h}}}$ is the $v$-th view's embedding (latent) representation of instance \textit{n}, which is viewed as view-specific representation, and we can reformulate it as ${{\bf{H}}^v} = \left\{ {{\bf{h}}_n^v} \right\}_{n = 1}^N \in {\mathbb{R}^{{M^h} \times N}}$.

\textbf{Consistent attentive layer:} we try to explore the consistent feature among multiple views by utilizing the following shared weights operation:
\begin{equation}
{\bf{h}}_{c,n}^v = {{\bf{W}}_c}\left[ {{\bf{h}}_n^v;1} \right],
\label{eq2}
\end{equation}
\begin{equation}
{{\bf{h}}_{c,n}} = \sum\nolimits_{v = 1}^V {\alpha _{c,n}^v{\bf{h}}_{c,n}^v},
\label{eq3} 
\end{equation}
where the weight $\alpha _c^v$, which denotes the contribution for the $v$-th view, is calculated according to attention mechanism \cite{NIPS2017_7181}:
\textcolor{black}{\begin{equation}
\begin{array}{l}
a_{c,n}^v = {\bf{q}}_c^{\rm T}\tanh \left( {{\bf{K}}_c^v\left[ {{\bf{h}}_{c,n}^v;1} \right]} \right),
\end{array}
\label{eq4}
\end{equation}}
\begin{equation}
\alpha _{c,n}^v = \frac{{\exp \left( {a_{c,n}^v} \right)}}{{\sum\nolimits_{p = 1}^V {\exp \left( {a_{c,n}^p} \right)} }}.
\label{eq5}
\end{equation}
Equation (\ref{eq2}) implement the process that project each view's embedding vector ${\bf{h}}_{n}^v$ onto a same subspace by using a shared weight ${\bf{W}}_{c}  \in {\mathbb{R}^{{M^h} \times \left( {{M^h} + 1} \right)}}$, and thus the $v$-th view's consistent representation is ${\bf{H}}_c^v = \left\{ {{\bf{h}}_{c,n}^v} \right\}_{n = 1}^N \in {\mathbb{R}^{{M^h} \times N}}$. Then we fuse multiple views' representation according equation (\ref{eq3}), and the updating rule of $\alpha _{c,n}^v$ is a special kind of attention mechanism \cite{Wu:2019:NDP:3289600.3291034} shown in equation (\ref{eq4}) and (\ref{eq5}). Unlike normal alignment operation, we only need one query ${\bf{q}}_c \in {\mathbb{R}^{{M^h}}}$ and corresponding keys ${\bf{K}}_c^v \in {\mathbb{R}^{{M^h}\times \left( {{M^h} + 1} \right)} }$ (they are just viewed as trainable variables) that we can compute $v$-th view's contribution weight $\alpha _c^v$ and obtain joint consistent representation ${{\bf{H}}_c} = \left\{ {{{\bf{h}}_{c,n}}} \right\}_{n = 1}^N \in {\mathbb{R}^{{M^h} \times N}}$ fused by $\left\{ \mathbf{H}_{c}^{v} \right\} _{v=1}^{V}$ according to weight $\left\{ \left\{ \alpha _{c,n}^{v} \right\} _{n=1}^{N} \right\} _{v=1}^{V}$.

\textbf{Global attentive layer:} through the above steps, we have obtained view-specific representations $\left\{ {{{\bf{H}}^1},{{\bf{H}}^2}, \cdots ,{{\bf{H}}^V}} \right\}$, and consistent representation ${\bf{H}}_c$, then we can fuse them to create a joint latent representation consists of complementary and consistent features acquired by following steps:
\begin{equation}
{{\bf{z}}_n} = \sum\nolimits_{v = 1}^V {{\alpha_n ^v}{\bf{h}}_n^v}  + {\alpha _{c,n}}{{\bf{h}}_{c,n}},
\label{eq6}
\end{equation}
\textcolor{black}{\begin{equation}
{a_n^v} = {\bf{q}^{\rm T}}\tanh \left( {{{\bf{K}}^v}\left[ {{\bf{h}}_n^v;1} \right]} \right),
\label{eq7}
\end{equation}}
\textcolor{black}{\begin{equation}
{a_{c,n}} = {\bf{q}^{\rm T}}\tanh \left( {{{\bf{K}}_c}\left[ {{{\bf{h}}_{c,n}};1} \right]} \right),
\label{eq8}
\end{equation}}
\begin{equation}
{\alpha_n^v} = \frac{{\exp \left( {{a_n^v}} \right)}}{{\sum\nolimits_{p = 1}^V {\exp \left( {{a_n^p}} \right) + \exp \left( {{a_{c,n}}} \right)} }},
\label{eq9}
\end{equation}
\begin{equation}
{\alpha _{c,n}} = \frac{{\exp \left( {{a_{c,n}}} \right)}}{{\sum\nolimits_{p = 1}^V {\exp \left( {{a_n^p}} \right) + \exp \left( {{a_{c,n}}} \right)} }}.
\label{eq10}
\end{equation}
Among them, ${\bf{Z}} = \left\{ {{{\bf{z}}_n}} \right\}_{n = 1}^N \in {\mathbb{R}^{{M^h} \times N}}$ is the joint latent representation fused by view-specific and consistent representations according to equation (\ref{eq6}) with the contribution weights $\left\{ \left\{ \alpha _{n}^{v} \right\} _{n=1}^{N} \right\} _{v=1}^{V}$ and $\left\{ \alpha _{c}^{n} \right\} _{n=1}^{N}$, and equation (\ref{eq7}) - (\ref{eq10}) are their update formulas; similar to consistent attentive layer, ${\bf{q}} \in {\mathbb{R}^{{M^h}}}$, ${\bf{K}}^v \in {\mathbb{R}^{{M^h}\times \left( {{M^h} + 1} \right)} }$, ${\bf{K}}_c \in {\mathbb{R}^{{M^h}\times \left( {{M^h} + 1} \right)} }$ represent query, view-specific keys, consistent key respectively.

\textbf{Self-representation layer:} the general self-representation subspace learning is usually formulated as follow:
\begin{equation*}
\begin{array}{l}
\mathop {\min }\limits_{\bf{C}} R\left( {\bf{C}} \right)\\
s.t.\;{\bf{X}} = {\bf{XC}},\;diag({\bf{C}}) = 0.
\end{array}
\label{eq11}
\end{equation*}
In this paper, we concern subspace clustering on joint latent representation $\bf{Z}$, therefore suppose that ${\bf{C}} \in {\mathbb{R}^{N \times N}}$ is the self-representation coefficient matrix and the process can be formulated as follows:
\begin{equation}
{{\bf{Z}}_s} = {\bf{ZC}},
\label{eq12}
\end{equation}
where ${{\bf{Z}}_s} = \left\{ {{{\bf{z}}_{s,n}}} \right\}_{n = 1}^N \in {\mathbb{R}^{{M^h} \times N}}$ is the reconstruction representation of $\bf{Z}$ through self-representation process, and the constraint on $\bf{C}$ will be discussed in the following subsections. Note that we cannot constraint the diagonal entries of $\bf{C}$ equal to zero in gradient descent, therefore when programming we can subtract the diagonal elements of $\bf{C}$ in advance before we implement equation (\ref{eq12}).

\textbf{Decoder layer:} we have obtained multiple views' global joint latent representation $\bf{Z}$ and its reconstruction representation ${\bf{Z}}_s$, suppose that ${\bf{Z}}_s$ can be utilized to reconstruct each view in reverse, and we have:
\begin{equation}
{\bf{\hat x}}_n^v = f\left( {{{{\bf{\bar W}}}_{\left( {L,v} \right)}}f\left( {{{{\bf{\bar W}}}_{\left( {L - 1,v} \right)}} \ldots f\left( {{{{\bf{\bar W}}}_{\left( {1,v} \right)}}\left[ {{{\bf{z}}_{s,n}};1} \right]} \right)} \right)} \right).
\label{eq13}
\end{equation}
Among of them, ${{\bf{\hat X}}^v} = \left\{ {{\bf{\hat x}}_n^v} \right\}_{n = 1}^N \in {\mathbb{R}^{{M^v} \times N}}$ is the reconstruction of the $v$-th view; ${{\bf{\bar W}}_{\left( {l,v} \right)}} \in {R^{{M^h} \times \left( {{M^h} + 1} \right)}}$ and ${{\bf{\bar W}}_{\left( {L,v} \right)}} \in {R^{{M^v} \times \left( {{M^h} + 1} \right)}}$ are the combination of fully connected layer's weights and biases, and $l \in \left\{ {1, \cdots ,L-1} \right\}$ stands for the \textit{l}-th layer of decoder; in addition, $f\left(  \cdot  \right)$ is the activation function, and in this paper we use Relu.

\textbf{Shortcut connection:} Based on discussion above, we have proposed a complicated multi-view framework. In practice, the layers number $L$ of encoder or decoder is usually set from 2 to 4, together with attentive layers and self-representation layer, the depth of our proposed model is relatively deep. Simultaneously, the main body of network structure is fully connected layer, which makes the model easy to degrade in training process. Fortunately, by introducing shortcut connection structure \cite{DBLP:journals/corr/HeZRS15}, we can effectively deal with the degradation problem and significantly improve learning performance. 

In this paper, we apply shortcut connection in the modules that have more than two layers, i.e., encoder, decoder, attentive layers, and therefore we should reformulate equation (\ref{eq1}), (\ref{eq6}), and (\ref{eq13}) respectively as follows:
\begin{equation*}
\begin{array}{l}
{\bf{h}}_n^v = {\bf{W}}_e^v{\bf{x}}_n^v + f\left( {{{\bf{W}}_{\left( {L,v} \right)}}f\left( {{{\bf{W}}_{\left( {L - 1,v} \right)}} \ldots f\left( {{{\bf{W}}_{\left( {1,v} \right)}}\left[ {{\bf{x}}_n^v;1} \right]} \right)} \right)} \right),
\end{array}
\label{eq14}
\end{equation*}

\begin{equation}
{{\bf{z}}_n} = \sum\nolimits_{v = 1}^V {{\alpha ^v}{\bf{h}}_n^v}  + {\alpha _c}{{\bf{h}}_{c,n}} + \frac{1}{V}\sum\nolimits_{v = 1}^V {{\bf{h}}_n^v},
\label{eq15}
\end{equation}

\begin{equation*}
\begin{array}{l}
{\bf{\hat x}}_n^v = {\bf{W}}_d^v{\bf{z}}_n + f\left( {{{{\bf{\bar W}}}_{\left( {L,v} \right)}}f\left( {{{{\bf{\bar W}}}_{\left( {L - 1,v} \right)}} \ldots f\left( {{{{\bf{\bar W}}}_{\left( {1,v} \right)}}\left[ {{{\bf{z}}_{s,n}};1} \right]} \right)} \right)} \right).
\end{array}
\label{eq16}
\end{equation*}
Among of them, ${\bf{W}}_e^v \in {\mathbb{R}^{{M^h} \times \left( {{M^V}} \right)}}$ and ${\bf{W}}_d^v \in {\mathbb{R}^{{M^V} \times \left( {{M^h}} \right)}}$ are linear projections to match the corresponding dimensions \cite{DBLP:journals/corr/HeZRS15}. Note that consistent attentive layer's depth is shallow, but this module associates with global attentive layer, which makes the framework complicated when they combined. As a consequence, we apply shortcut connection only in global attentive layer in equation (\ref{eq15}), and because there are multiple inputs in this sub-module, we need to calculate their mean value.

This module is significant for our proposed algorithm, because the model is a complex hybrid network and will degrade when training without it. The detail discussion on it is in subsection \ref{sec 4.6}.  

\subsection{Loss Function}
In this paper, we attempt to use deep learning framework to achieve subspace clustering, so the loss function mainly contains two parts. Based on the assumptions of subsection \ref{sec 2.2}, the first part of loss function is referred to auto-encoder reconstruction loss:
\textcolor{black}{\begin{equation}
\mathop {\min }\limits_{\bf{\Theta }} \frac{1}{NV}\sum\nolimits_{v = 1}^V {\left\| {{{\bf{X}}^v} - {{{\bf{\hat X}}}^v}} \right\|_F^2}  + \lambda \Omega \left( {{\bf{W}},{\bf{\bar W}},{{\bf{W}}_c}} \right),
\label{eq17}
\end{equation}
where ${\bf{\Theta }}=\{{\bf{C}},{\bf{W}},{\bf{\bar W}},{{\bf{W}}_c},{\bf{K}}^v,{\bf{K}}_c^v,{\bf{q}},{\bf{q}}_c\}$ is the variable parameters set}, $\Omega \left(  \cdot  \right)$ is the regularization term of model parameters, and $\lambda$ is the trade-off parameter. In this paper, two alternatives to regularization terms are $\textit{l}_1$-norm and $\textit{l}_2$-norm, which can be viewed as a hyper-parameter. Then, the second part of loss function is referred to self-representation subspace learning:
\textcolor{black}{\begin{equation}
\mathop {\min }\limits_{{\bf{C}},{\bf{W}},{{\bf{W}}_c},{\bf{K}}^v,{\bf{K}}_c^v,{\bf{q}},{\bf{q}}_c} R\left( {\bf{C}} \right) + \frac{{{\lambda}}}{N}\left\| {{\bf{Z}} - {{\bf{Z}}_s}} \right\|_F^2,
\label{eq18}
\end{equation}}
where $R\left(  \cdot  \right)$ is regularization term, and $\lambda$ is the trade-off parameter. In subspace learning, three alternatives to regularization terms are usually $\textit{l}_1$-norm, nuclear norm, and frobenius norm, which correspond to sparse \cite{6482137}, low-rank \cite{Vidal2014}, least squares regression \cite{10.1007/978-3-642-33786-4_26} subspace learning respectively. In this paper, to facilitate optimization, we select frobenius norm as the regularization term, i.e., equation (\ref{eq18}) can be reformulated as:
\textcolor{black}{\begin{equation}
\mathop {\min }\limits_{{\bf{C}},{\bf{W}},{{\bf{W}}_c},{\bf{K}}^v,{\bf{K}}_c^v,{\bf{q}},{\bf{q}}_c} \left\| {\bf{C}} \right\|_F^2 + \frac{{{\lambda}}}{N}\left\| {{\bf{Z}} - {{\bf{Z}}_s}} \right\|_F^2.
\label{eq19}
\end{equation}}
In summary, we combine equation (\ref{eq17}) and (\ref{eq19}) to formulate joint object function:
\textcolor{black}{\begin{equation}
\begin{array}{l}
\mathop {\min }\limits_{\bf{\Theta }} \left\| {\bf{C}} \right\|_F^2 + \frac{{{\lambda _1}}}{N}\left\| {{\bf{Z}} - {{\bf{Z}}_s}} \right\|_F^2\\
+ {\lambda _2}\left[ {\frac{1}{NV}\sum\nolimits_{v = 1}^V {\left\| {{{\bf{X}}^v} - {{{\bf{\hat X}}}^v}} \right\|_F^2}  + {\lambda _3}\Omega \left( {{\bf{W}},{\bf{\bar W}},{{\bf{W}}_c}} \right)} \right]
\end{array},
\label{eq20}
\end{equation}}
where ${\lambda _1}$, ${\lambda _2}$, and ${\lambda _3}$ are trade-off parameters. As a consequence, by minimizing equation (\ref{eq20}) with stochastic gradient descent, we can easily optimize self-representation matrix $\bf{C}$. In other words, we only need to build forward process of AMVDSN using deep learning framework like Tensorflow and Pytorch when programming without considering the specific optimization process.

\textcolor{black}{\subsection{Pretraining Strategy}
Our proposed AMVDSN is a complex hybrid network, which is hard to directly train from scratch and trivial all-zero solution may appear when minimizing the loss function. A very effective and simple solution is to introduce pretraining strategy. We first train an auto-encoder (without attention, shortcut connection, and self-presentation process) for each view separately to prevent interference from other views. The process is helpful for quickly finding each view's view-specific embedding representation, multiple views encoders' weights construct ${\bf{W}}_0$, and decoders' weights are used to construct ${\bf{\bar{W}}}_0$. We then set the initial values of AMVDSN's weights ${\bf{W}}$ and ${\bf{\bar{W}}}$ as ${\bf{W}}_0$ and ${\bf{\bar{W}}}_0$, and train AMVDSN according to equation (\ref{eq20}) with fixed ${\bf{W}}$.} 

\textcolor{black}{Note that we fix ${\bf{W}}$ because the embedding is good enough to represents the corresponding view. However, ${\bf{\bar{W}}}$ is need to update because the input of AMVDSN's decoders is ${\bf{Z}}_s$ as shown in equation (\ref{eq13}), which is different from the inputs of pretraining stage's decoders (in pretraining stage, the input of $v$-th view's decoder is ${\bf{H}}^v$). The training process is simple, the configuration is same as the corresponding modules in AMVDSN, and with an early stopping constraint (loss no longer decreases within 200 epochs), each view's net usually converges in hundreds to thousands of epochs.}

\textcolor{black}{The pretraining based method is named as AMVDSN-ft, extensive experiments indicate that with the strategy can significantly improve AMVDSN's performance on most data sets, and the detailed experiments will be discussed in section \ref{sec 4}.}

\section{Experiment}
\label{sec 4}
To verify the performance of our proposed method, we compare AMVDSN with the 5 baseline methods on seven real-world data sets, where one method has additional 3 variants, i.e., there are actually 8 comparison methods. In this section, we will first introduce the data sets and baseline methods, and then discuss the experiments we designed.

\subsection{Data Sets}
In this subsection, we will introduce the data construction method for short, and seven data sets include ORL$\footnote{http://www.cad.zju.edu.cn/home/dengcai/Data/FaceData.html \label{f1}}$, Reuters \cite{10.5555/1005332.1005345}, 3-sources$\footnote{http://mlg.ucd.ie/datasets/3sources.html}$, Yale$^{\ref{f1}}$, uci-digit$\footnote{http://archive.ics.uci.edu/ml/datasets/Multiple+Features}$, Prokaryotic \cite{10.1093/nar/gkw964}, and NUS-WIDE-OBJ $\footnote{https://lms.comp.nus.edu.sg/wp-content/uploads/2019/research/nuswide/NUS-WIDE.html}$.

 \textbf{ORL}: a face data with 400 images of 40 subjects, which means each category involves only 10 pictures, and three types of features (intensity, LBP and Gabor) are utilized to construct three views. 
 
 \textbf{Yale}: a face data includes 15 different persons, and there are 11 facial expression or configuration for each category. Similar to ORL, three types of features (intensity, LBP and Gabor) are utilized to construct three views.
 
 \textbf{uci-digit}: a data set of handwritten digit with 10 categories (0-9), and each digit has 200 samples. The data contains 6 types of features, where 76 Fourier coefficients of the character shapes, 216 profile correlations, and 64 Karhunen-Love coefficients are selected to construct three views. 
 
 \textbf{Reuters}: a documents data with 6 categories which originally written in English, and they are also translated into French, German, Spanish and Italian. We view each language as a view therefore we obtain five views. All documents are in the bag-of-words representations, and we sampled 100 documents from each category to build the data set.
 
 \textbf{3-sources}: a news data set, where 169 news belong to some dominant topic classes are available in three different online news sources (BBC, Reuters, and The Guardian), the report from each source is viewed as a view. All articles are in the bag-of-words representations.
 
 \textbf{Prokaryotic}: a data set describes 551 prokaryotic species with heterogeneous representations, the first is instance's textual description, which is in the bag-of-words representation and is viewed as the first view. And another two genmic representations: 1) the proteome composition, encoded as relative frequencies of amino acids; 2) the gene repertoire, encoded as presence/absence indicators of gene families in a genome, are viewed as the remaining two views. Then component analysis (PCA) is utilized to reduce the dimensionality on three views respectively, and note that we retain principal components explaining 90\% of the variance.
 
 \textcolor{black}{\textbf{NUS-WIDE-OBJ}: a subset of NUS-WIDE data set with 30000 samples of 31 objects, which is intended for several object-based tasks. The data involves six types of features, and we select five of them (color histogram, color correlogram, edge direction histogram, wavelet texture, and block-wise color moments) as five views. Since the highly computational complexity of most compared baseline methods, we randomly select 1500 samples from raw data, and it is turn out that there are not too many differences between the clustering performance on the subset and whole set.} 
 
 In summary, the detail information of each data are summarized in Table \ref{table 1}, where $V$, $C$, $N$ and $M^v$ represent views number, clusters number, instance number and each view's dimension respectively.

\begin{table}[h!]
	\fontsize{7}{9.5}\selectfont
	\setlength{\abovecaptionskip}{0cm} 
	\setlength{\belowcaptionskip}{-0.2cm}
	\caption{Description of Data Sets}
	\label{table 1}
	\begin{center} 
		\begin{tabular}{ccccc}
			\hline
			Data        & $V$ & $C$ & $N$  & $M^v$                                                                                                 \\ \hline
			ORL          & 3   & 40  & 400   & {[}4096, 3304, 6750{]}                                                                                \\ \hline
			Reuters      & 5   & 6   & 600   & \multicolumn{1}{l}{\begin{tabular}[c]{@{}l@{}}{[}21526, 24892, 34121, 15487, 11539{]}\end{tabular}}   \\ \hline
			3-sources    & 3   & 6   & 169   & {[}3560, 3631, 3068{]}                                                                                \\ \hline
			Yale         & 3   & 15  & 165   & {[}4096, 3304, 6750{]}                                                                                \\ \hline
			uci-digit    & 3   & 10  & 2000  & {[}216, 76, 64{]}                                                                                     \\ \hline
			Prokaryotic  & 3   & 4   & 551   & {[}438, 3, 393{]}                                                                                     \\ \hline
			NUS-WIDE-OBJ & 5   & 31  & 30000 & {[}64, 225, 144, 73, 128{]}                                                                           \\ \hline
		\end{tabular}
	\end{center}
\end{table}

\subsection{Baseline Methods}
\label{sec 4.2}
To demonstrate the efficiency of our proposed algorithm, some state-of-the-art subspace clustering algorithms are selected as the baseline methods:

\textbf{LRSC} \cite{Vidal2014}:  a single view subspace clustering method named Low-Rank Subspace Clustering, we conduct the algorithm on all views and select the best result.

\textbf{LMSC} \cite{8099944}: a multi-view subspace clustering method named latent Multi-view Subspace Clustering. The algorithm is based on the assumption that all views come from a same latent representation through projection of multiple certain orthogonal matrix. And the algorithm applys the latent representation into self-representation subspace clustering with nuclear norm ruglarization.

\textbf{CSMSC} \cite{Luo2018}: a multi-view subspace clustering method named Consistent and Specific Multi-View Subspace Clustering. The algorithm is based on the assumption that self-representation matrix includes consistent and specific parts, and uses different norms to regularize them, where nuclear norm and frobenius norm are consistent and specific parts' regularization terms respectively.

\textbf{MLRSSC} \cite{Brbic2018}: a multi-view subspace clustering method named Multi-view Low-Rank Sparse Subspace Clustering, which let the self-representation matrix sparse and low-rank simultaneously. Besides, the paper provides two alignment methods to ensure the multiple views' consistent property, the first let each view adjacent to the rest views, whcih is also MLRSSC's idea; the second let all views adjacent to a center (a new variable), which is called centroid-based multi-view low-rank sparse subspace clustering (\textbf{MLRSSC-C}). On the other hand, to deal with non-linear data, the paper proposed kernel style variants based on two aforementioned assumptions, which are abbreviated as \textbf{KMLRSSC} and \textbf{MLRSSC-C}.

\textcolor{black}{\textbf{MvDSCN} \cite{DBLP:journals/corr/abs-1908-01978}: a neural network based framework named Multi-view Deep Subspace Clustering Networks, which aims to discover the inherent structure by fusing multi-view complementary information. Up until now, only the source code on a multi-modal data set has been released, therefore we only compare with this algorithm on ORL and Yale data sets because MvDSCN was also conducted comparison experiments on them in reference \cite{DBLP:journals/corr/abs-1908-01978}.} 

\textcolor{black}{\textbf{CSI} \cite{Weng2020}: a multi-view subspace clustering method named Common Subspace Integration. The algorithm conducts self-representation reconstruction on each view and integrates different subspace into common subspace using a graph-based method. Compare with most multi-view subspace algorithms that just calculate the mean of multiple subspace, CSI integrates subspace through learning and works well. Although it gains performance enhancements, but also causes higher computational complexity.}

\textcolor{black}{\textbf{FCMSC} \cite{Zheng2020}: a multi-view subspace clustering method named Feature Concatenation Multi-view Subspace Clustering. The algorithm deals with the problem that multiple views have different statistic properties and simply concatenating them directly cannot derive a satisfied clustering performance.}  

\begin{table}[h!]
	\setlength{\abovecaptionskip}{0cm} 
	\setlength{\belowcaptionskip}{-0.2cm}
	\fontsize{7}{9.5}\selectfont
	\caption{Model Configuration}
	\label{table 2}
	\begin{center} 
		\begin{tabular}{cccc}
			\hline
			Data         & Layers Setting                                                    & $\lambda_1$, $\lambda_2$, $\lambda_3$ & Regularization \\ \hline
			ORL          & \begin{tabular}[c]{@{}c@{}}{[}128, 128, 128, 128{]}\end{tabular}  & {[}0.5, 0.5, 0.1{]}                     & $l_2$-norm         \\ \hline
			Reuters      & {[}512, 512{]}                                                    & {[}0.5, 0.5, 0.01{]}                    & $l_2$-norm         \\ \hline
			3-sources    & {[}512, 512{]}                                                    & {[}0.5, 0.5, 0.1{]}                     & $l_1$-norm         \\ \hline
			Yale         & {[}128, 128{]}                                                    & {[}0.5, 0.5, 0.1{]}                     & $l_2$-norm         \\ \hline
			uci-digit    & {[}512, 512{]}                                                    & {[}0.5, 0.5, 0.1{]}                     & $l_1$-norm         \\ \hline
			Prokaryotic  & {[}128, 128{]}                                                    & {[}0.5, 0.5, 0.1{]}                     & $l_1$-norm         \\ \hline
			NUS-WIDE-OBJ & {[}256, 256, 256{]}                                               & {[}0.5, 0.5, 0.01{]}                    & $l_2$-norm        \\ \hline
		\end{tabular}
	\end{center}
\end{table}

\begin{table*}[h!]
	\fontsize{7}{9.5}\selectfont
	\setlength{\abovecaptionskip}{0cm} 
	\setlength{\belowcaptionskip}{-0.2cm}
	\caption{Clustering results on ORL, Prokaryotic, and Yale.}
	\label{table 3}
	\begin{center}
		\begin{tabular}{c|c|cccccc}
			Dataset                & Method           & ACC                   & NMI                   & ARI                    & Precision             & Recall                & F-score               \\ \hline
			\multirow{12}{*}{ORL}  & LRSC             & 0.769(0.015)          & 0.883(0.015)          & 0.688(0.037)          & 0.668(0.037)          & 0.726(0.036)          & 0.695(0.036)          \\
			& LMSC             & 0.822(0.037)          & 0.927(0.013)          & 0.774(0.043)          & 0.813(0.038)          & 0.822(0.037)          & 0.802(0.041)          \\
			& CSMSC            & 0.868(0.0012)         & 0.942(0.005)          & 0.827(0.002)          & 0.860(0.002)          & 0.804(0.003)          & 0.831(0.001)          \\
			& MLRSSC           & 0.637(0.034)          & 0.813(0.017)          & 0.524(0.037)          & 0.640(0.033)          & 0.637(0.034)          & 0.623(0.034)          \\
			& MLRSSC-Centroid  & 0.780(0.027)          & 0.917(0.009)          & 0.729(0.029)          & 0.785(0.033)          & 0.780(0.027)          & 0.761(0.032)          \\
			& KMLRSSC          & 0.786(0.041)          & 0.903(0.016)          & 0.721(0.042)          & 0.793(0.040)          & 0.786(0.041)          & 0.774(0.042)          \\
			& KMLRSSC-Centroid & 0.783(0.031)          & 0.907(0.008)          & 0.721(0.026)          & 0.793(0.039)          & 0.783(0.031)          & 0.772(0.035)          \\
			& CSI              & 0.863(0.023)          & 0.930(0.009)          & 0.805(0.025)          & 0.782(0.029)          & 0.840(0.020)          & 0.810(0.025)          \\
			& FCSMC            & 0.847(0.019)          & 0.928(0.008)          & 0.782(0.021)          & 0.746(0.025)          & 0.833(0.020)          & 0.787(0.020)          \\
			& MvDSCN           & 0.870(0.006)          & 0.943(0.002)          & 0.819(0.001)          & \textbackslash{}      & \textbackslash{}      & 0.834(0.012)          \\
			& AMVSC            & 0.887(0.010)          & 0.936(0.003)          & 0.831(0.021)          & 0.885(0.022)          & 0.887(0.010)          & 0.887(0.013)          \\
			& AMVSC-ft         & \textbf{0.921(0.005)} & \textbf{0.958(0.003)} & \textbf{0.882(0.007)} & \textbf{0.918(0.005)} & \textbf{0.921(0.005)} & \textbf{0.914(0.005)} \\ \hline
			\multirow{11}{*}{Prokaryotic} & LRSC             & 0.582(0.005)          & 0.079(0.011)          & 0.038(0.029)          & 0.407(0.011)          & \textbf{0.964(0.047)} & 0.571(0.001)          \\
			& LMSC             & 0.709(0.032)          & 0.418(0.034)          & 0.394(0.052)          & 0.669(0.067)          & 0.585(0.068)          & 0.618(0.026)          \\
			& CSMSC            & 0.658(0.006)          & 0.352(0.004)          & 0.367(0.005)          & 0.665(0.005)          & 0.527(0.007)          & 0.588(0.004)          \\
			& MLRSSC           & 0.646(0.038)          & 0.309(0.018)          & 0.324(0.037)          & 0.529(0.011)          & 0.531(0.006)          & 0.498(0.005)          \\
			& MLRSSC-Centroid  & 0.620(0.009)          & 0.196(0.008)          & 0.260(0.005)          & 0.444(0.033)          & 0.399(0.018)          & 0.363(0.024)          \\
			& KMLRSSC          & 0.638(0.045)          & 0.420(0.041)          & 0.355(0.071)          & 0.609(0.032)          & 0.641(0.053)          & 0.589(0.041)          \\
			& KMLRSSC-Centroid & 0.655(0.045)          & 0.405(0.030)          & 0.350(0.072)          & 0.606(0.031)          & 0.657(0.046)          & 0.597(0.042)          \\
			& CSI              & 0.619(0.000)          & 0.413(0.000)          & 0.291(0.000)          & 0.634(0.000)          & 0.440(0.000)          & 0.520(0.000)          \\
			& FCSMC            & 0.674(0.001)          & 0.443(0.000)          & 0.394(0.001)          & 0.688(0.001)          & 0.537(0.001)          & 0.603(0.001)          \\
			& AMVSC            & 0.823(0.301)          & 0.543(0.265)          & 0.593(0.883)          & 0.758(0.651)          & 0.794(0.744)          & 0.762(0.733)          \\
			& AMVSC-ft         & \textbf{0.872(0.023)} & \textbf{0.592(0.033)} & \textbf{0.683(0.047)} & \textbf{0.827(0.445)} & 0.861(0.009)          & \textbf{0.837(0.034)} \\ \hline
			\multirow{12}{*}{Yale} & LRSC             & 0.675(0.027)          & 0.706(0.019)          & 0.491(0.032)          & 0.501(0.030)          & 0.547(0.030)          & 0.523(0.030)          \\
			& LMSC             & 0.789(0.018)          & 0.789(0.019)          & 0.628(0.030)          & 0.632(0.029)          & 0.672(0.027)          & 0.651(0.028)          \\
			& CSMSC            & 0.752(0.001)          & 0.784(0.001)          & 0.615(0.005)          & 0.673(0.002)          & 0.610(0.006)          & 0.794(0.029)          \\
			& MLRSSC           & 0.660(0.050)          & 0.697(0.033)          & 0.477(0.054)          & 0.482(0.056)          & 0.544(0.045)          & 0.511(0.050)          \\
			& MLRSSC-Centroid  & 0.627(0.034)          & 0.702(0.021)          & 0.458(0.024)          & 0.711(0.050)          & 0.627(0.034)          & 0.648(0.041)          \\
			& KMLRSSC          & 0.649(0.053)          & 0.689(0.032)          & 0.485(0.046)          & 0.679(0.070)          & 0.649(0.053)          & 0.653(0.059)          \\
			& KMLRSSC-Centroid & 0.639(0.043)          & 0.680(0.032)          & 0.480(0.044)          & 0.661(0.046)          & 0.639(0.043)          & 0.640(0.045)          \\
			& CSI              & 0.734(0.000)          & 0.777(0.000)          & 0.606(0.000)          & 0.606(0.000)          & 0.659(0.000)          & 0.739(0.000)          \\
			& FCSMC            & 0.775(0.025)          & 0.796(0.019)          & 0.589(0.041)          & 0.569(0.046)          & 0.674(0.028)          & 0.617(0.037)          \\
			& MvDSCN           & 0.824(0.004)          & 0.797(0.007)          & 0.626(0.011)          & \textbackslash{}      & \textbackslash{}      & 0.650(0.010)          \\
			& AMVSC            & 0.781(0.022)          & 0.747(0.011)          & 0.540(0.074)          & 0.870(0.063)          & 0.782(0.022)          & 0.806(0.022)          \\
			& AMVSC-ft         & \textbf{0.867(0.012)} & \textbf{0.844(0.013)} & \textbf{0.685(0.047)} & \textbf{0.921(0.021)} & \textbf{0.867(0.012)} & \textbf{0.877(0.009)}
		\end{tabular}
	\end{center}
\end{table*}

\subsection{Model Configuration}
Different data sets require different model configurations, and firstly, the initial method of $\bf{W}$, $\bf{\bar W}$ and ${\bf{W}}_c$ is Lecun normal initializer \cite{lecun1998efficient,NIPS2017_6698}, which is a function in keras: keras.initializers.lecun\_norm. For optimizer, Adam \cite{DBLP:journals/corr/KingmaB14} with the learning rate 0.001 is satisfied, and usually after 100 - 200 epochs, the network converge to optimal results. Besides, the setting of each layer's node number, regularization, trade-off parameters $\lambda_1$, $\lambda_2$, and $\lambda_3$ are summarized in Table \ref{table 2}.

\subsection{Experimental Setting}
Before we formally discuss the experimental results, we need to explain some experimental considerations. 

(1) Once the network is trained, we can use the self-representation layer's weight $\bf{C}$ to construct affinity matrix of spectral clustering, and the calculation rule is:
\begin{equation*}
\frac{\left| \mathbf{C} \right|+\left| \mathbf{C}^{\text{T}} \right|}{2}.
\end{equation*}
 However, over the years, researchers have developed some heuristics to calculate better affinity matrix, especially for neural network based model, the constraint on $\bf{C}$ is not strict,  therefore we utilize the method proposed by \cite{6836065} to compute the affinity matrix of the spectral clustering will derive better learning performance.
 
 (2) We select 6 metrics to measure the clustering effect, and they are NMI (Normalized Mutual Information), ARI (Adjusted Rand Index), ACC (Accuracy), Precision, Recall, and F-score. The first two ones are traditional clustering metrics,  but the last four ones are actually classification metrics, therefore we need to compute the best map of the predicted labels compared with the ground truth using Kuhn-Munkres algorithm first. Note that higher values indicates better performance for all our selected metrics.

\begin{table*}[h!]
	\fontsize{7}{9.5}\selectfont
	\setlength{\abovecaptionskip}{0cm} 
	\setlength{\belowcaptionskip}{-0.2cm}
	\caption{Clustering results on Reuters, 3-sources, uci-digit and NUS-WIDE-OBJ.}
	\label{table 4}
	\begin{center} 
		\begin{tabular}{c|ccccccc}
			Dataset                       & Method           & ACC                   & NMI                   & ARI                    & Precision             & Recall                & F-score               \\ \hline
			\multirow{11}{*}{Reuters}     & LRSC             & 0.209(0.007)          & 0.131(0.007)          & 0.015(0.001)          & 0.172(0.000)          & \textbf{0.835(0.013)} & 0.285(0.001)          \\
			& LMSC             & 0.495(0.001)          & 0.355(0.001)          & 0.223(0.010)          & 0.312(0.009)          & 0.497(0.008)          & 0.383(0.006)          \\
			& CSMSC            & 0.513(0.001)          & 0.358(0.001)          & 0.247(0.001)          & 0.343(0.001)          & 0.444(0.001)          & 0.387(0.001)          \\
			& MLRSSC           & 0.530(0.033)          & 0.382(0.015)          & 0.282(0.026)          & 0.501(0.055)          & 0.530(0.033)          & 0.483(0.051)          \\
			& MLRSSC-Centroid  & 0.514(0.035)          & 0.370(0.011)          & 0.268(0.026)          & 0.484(0.040)          & 0.514(0.035)          & 0.459(0.046)          \\
			& KMLRSSC          & 0.571(0.023)          & 0.374(0.019)          & 0.304(0.020)          & 0.618(0.037)          & 0.571(0.023)          & 0.567(0.030)          \\
			& KMLRSSC-Centroid & 0.551(0.024)          & 0.357(0.016)          & 0.294(0.016)          & 0.591(0.046)          & 0.551(0.024)          & 0.540(0.027)          \\
			& CSI              & 0.527(0.009)          & 0.394(0.009)          & 0.288(0.007)          & 0.364(0.006)          & 0.511(0.008)          & 0.425(0.007)          \\
			& FCSMC            & 0.518(0.001)          & 0.369(0.001)          & 0.268(0.000)          & 0.357(0.000)          & 0.470(0.001)          & 0.405(0.000)          \\
			& AMVSC            & 0.592(0.005)          & 0.396(0.006)          & 0.315(0.006)          & 0.642(0.005)          & 0.592(0.005)          & 0.583(0.006)          \\
			& AMVSC-ft         & \textbf{0.650(0.006)} & \textbf{0.459(0.009)} & \textbf{0.380(0.010)} & \textbf{0.690(0.006)} & 0.650(0.006)          & \textbf{0.648(0.006)} \\ \hline
			\multirow{11}{*}{3-sources}   & LRSC             & 0.596(0.026)          & 0.462(0.009)          & 0.351(0.020)          & 0.471(0.037)          & 0.585(0.054)          & 0.518(0.004)          \\
			& LMSC             & 0.734(0.020)          & 0.676(0.018)          & 0.579(0.017)          & 0.729(0.024)          & 0.621(0.047)          & 0.669(0.018)          \\
			& CSMSC            & 0.798(0.002)          & 0.747(0.005)          & 0.673(0.004)          & 0.730(0.005)          & 0.773(0.001)          & 0.751(0.003)          \\
			& MLRSSC           & 0.677(0.060)          & 0.593(0.026)          & 0.550(0.067)          & 0.535(0.048)          & 0.596(0.060)          & 0.544(0.059)          \\
			& MLRSSC-Centroid  & 0.666(0.051)          & 0.590(0.019)          & 0.534(0.053)          & 0.512(0.045)          & 0.551(0.068)          & 0.513(0.056)          \\
			& KMLRSSC          & 0.609(0.038)          & 0.526(0.027)          & 0.423(0.035)          & 0.536(0.59)           & 0.564(0.079)          & 0.517(0.060)          \\
			& KMLRSSC-Centroid & 0.621(0.025)          & 0.531(0.020)          & 0.457(0.031)          & 0.544(0.057)          & 0.562(0.051)          & 0.519(0.046)          \\
			& CSI              & 0.763(0.000)          & 0.747(0.000)          & 0.622(0.000)          & 0.717(0.000)          & 0.702(0.000)          & 0.709(0.000)          \\
			& FCSMC            & \textbf{0.908(0.003)} & \textbf{0.795(0.006)} & \textbf{0.793(0.008)} & \textbf{0.887(0.004)} & \textbf{0.795(0.007)} & \textbf{0.839(0.006)} \\
			& AMVSC            & 0.823(0.014)          & 0.749(0.024)          & 0.735(0.054)          & 0.685(0.019)          & 0.758(0.006)          & 0.710(0.011)          \\
			& AMVSC-ft         & 0.831(0.010)          & 0.741(0.016)          & 0.718(0.020)          & 0.727(0.039)          & 0.791(0.033)          & 0.746(0.035)          \\ \hline
			\multirow{11}{*}{uci-digit}   & LRSC             & 0.763(0.003)          & 0.697(0.001)          & 0.626(0.003)          & 0.650(0.003)          & 0.678(0.002)          & 0.664(0.003)          \\
			& LMSC             & 0.905(0.004)          & 0.833(0.006)          & 0.812(0.008)          & 0.835(0.007)          & 0.840(0.007)          & 0838(0.007)           \\
			& CSMSC            & 0.903(0.000)          & 0.835(0.001)          & 0.810(0.000)          & 0.823(0.000)          & 0.834(0.004)          & 0.828(0.000)          \\
			& MLRSSC           & 0.881(0.061)          & 0.852(0.022)          & 0.814(0.050)          & 0.877(0.070)          & 0.881(0.061)          & 0.873(0.070)          \\
			& MLRSSC-Centroid  & 0.893(0.055)          & 0.853(0.023)          & 0.818(0.050)          & 0.891(0.063)          & 0.893(0.055)          & 0.887(0.063)          \\
			& KMLRSSC          & 0.894(0.049)          & 0.861(0.018)          & 0.826(0.044)          & 0.888(0.060)          & 0.894(0.049)          & 0.886(0.058)          \\
			& KMLRSSC-Centroid & 0.910(0.046)          & 0.865(0.018)          & 0.840(0.039)          & 0.910(0.051)          & 0.910(0.046)          & 0.906(0.053)          \\
			& CSI              & 0.863(0.000)          & \textbf{0.915(0.000)} & 0.846(0.000)          & 0.805(0.000)          & 0.929(0.000)          & 0.862(0.000)          \\
			& FCSMC            & 0.916(0.001)          & 0.848(0.001)          & 0.827(0.001)          & 0.840(0.001)          & 0.849(0.001)          & 0.844(0.001)          \\
			& AMVSC            & 0.932(0.001)          & 0.876(0.005)          & 0.859(0.004)          & 0.935(0.001)          & 0.932(0.001)          & 0.932(0.001)          \\
			& AMVSC-ft         & \textbf{0.958(0.001)} & 0.909(0.002)          & \textbf{0.909(0.002)} & \textbf{0.958(0.001)} & \textbf{0.958(0.001)} & \textbf{0.958(0.001)} \\ \hline
			\multirow{11}{*}{NUS-WIDE-OBJ}  & LRSC             & 0.139(0.002)          & 0.174(0.002)          & 0.027(0.002)          & 0.097(0.002)          & 0.055(0.001)          & 0.070(0.001)          \\
			& LMSC             & 0.177(0.004)          & 0.211(0.004)          & 0.056(0.003)          & 0.138(0.003)          & 0.074(0.002)          & 0.096(0.003)          \\
			& CSMSC            & 0.169(0.004)          & 0.213(0.003)          & 0.050(0.003)          & 0.128(0.004)          & 0.071(0.002)          & 0.091(0.002)          \\
			& MLRSSC           & 0.160(0.007)          & 0.209(0.005)          & 0.047(0.003)          & 0.125(0.004)          & 0.067(0.002)          & 0.088(0.003)          \\
			& MLRSSC-Centroid  & 0.158(0.010)          & 0.205(0.006)          & 0.044(0.005)          & 0.121(0.006)          & 0.065(0.003)          & 0.085(0.004)          \\
			& KMLRSSC          & 0.173(0.006)          & 0.183(0.004)          & 0.043(0.001)          & 0.095(0.002)          & 0.146(0.017)          & 0.114(0.005)          \\
			& KMLRSSC-Centroid & 0.168(0.008)          & 0.204(0.006)          & 0.048(0.004)          & 0.125(0.005)          & 0.070(0.003)          & 0.090(0.004)          \\
			& CSI              & 0.171(0.002)          & 0.167(0.003)          & 0.032(0.002)          & 0.086(0.002)          & 0.137(0.001)          & 0.105(0.001)          \\
			& FCSMC            & 0.149(0.003)          & 0.188(0.003)          & 0.039(0.001)          & 0.115(0.002)          & 0.062(0.001)          & 0.080(0.001)          \\
			& AMVSC            & \textbf{0.197(0.003)} & \textbf{0.229(0.004)} & \textbf{0.067(0.004)} & \textbf{0.178(0.006)} & \textbf{0.188(0.008)} & \textbf{0.156(0.004)} \\
			& AMVSC-ft         & 0.193(0.002)          & 0.227(0.006)          & 0.065(0.004)          & 0.176(0.006)          & 0.184(0.009)          & 0.153(0.003)         
		\end{tabular}
	\end{center} 
\end{table*}

\subsection{Clustering Result Analysis}

We conduct the experiments on seven real-world data sets using 10 baseline methods and our proposed AMVDSN and AMVDSN-ft, and then summarize the clustering results in Table \ref{table 3} and \ref{table 4}. Obviously, AMVDSN-ft have the best performance on most data sets, especially on Prokaryotic, and it has improved performance by more than 10 percentage points on all metrics; we also find even if the pretraining strategy is not used, AMVDSN is very stable on all data sets, and achieve the best performance on NUS-WIDE-OBJ.

Experimental results indicate that compare with traditional matrix factorization based methods, AMVDSN finds better representation so that despite self-representation matrix is obtained through a simple strategy (the strategy is formulated as equation (\ref{eq19})), we still get much better clustering results. For example, \cite{6482137,Vidal2014,Luo2018} don't have additional feature extraction process; \cite{8099944} computes a latent representation of multiple views but the process is actually a linear transformation; the kernel style variants of \cite{Brbic2018} need first map the original input into a high dimensional feature space. However, their strategies perform not well, and compare with them, our proposed method is based on neural network which has good non-linear ability, and attention provides AMVDSN with better views fusion performance. 

\textcolor{black}{On the other hand, compare with multi-view deep learning method MvDSCN we still have two advantages: (1) AMVSC-ft performs better than MvDSCN on ORL and Yale (we only list the results on such two data sets, and reason is discussed in subsection \ref{sec 4.2}); (2) our pretraining strategy is much easier to train than MvDSCN. In pretraining stage, MvDSCN train multiple auto-encoders simultaneously, the loss function is all net's joint reconstruction error, which converges slowly and often not converges after tens of thousands of epochs. But our strategy is to train multiple views' auto-encoders respectively, with a simple early stopping method (loss no longer decreases within 200 rounds), each view's auto-encoder usually converges in only hundreds to thousands of epochs.}

\begin{figure}[h!]
	\begin{center}
		\includegraphics[width=0.4\textwidth]{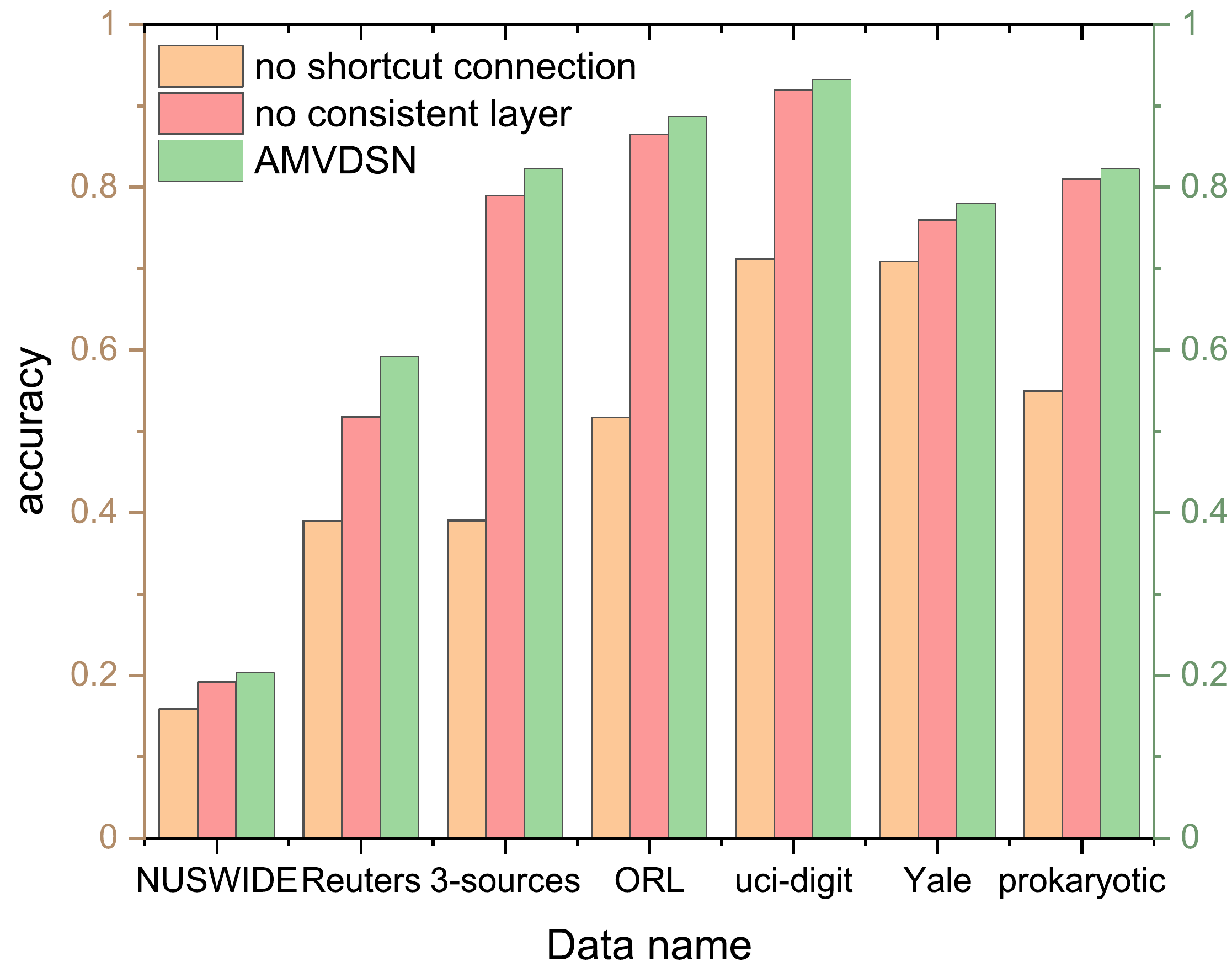}
		\caption{Ablation Study: the orange, red, and green bars show the clustering performance of three types of AMVDSN: no shortcut connection, no consistent attentive layers, and the standard AMVDSN.} \label{figure_0}
	\end{center}
\end{figure}

\textcolor{black}{\subsection{Ablation Study}
\label{sec 4.6}
In this subsection, we design some ablation experiments to verify the roles of shortcut connection and consistent attentive layer modules. AMVDSN is set as the baseline, we remove shortcut connection layers and consistent attentive layers respectively and compare their clustering accuracy, and the comparisons of them are shown in Figure \ref{figure_0}. The orange bars indicate that training from scratch without the shortcut connection module is hard or even infeasible on some data set, the reason is AMVDSN's structure is deep and complex such that model degrades when training. Fortunately, shortcut connection deals with the problem well and even makes AMVDSN could train from scratch. When using residual module and remove consistent layer, the red bars indicate that the model can be well-trained, but the clustering performance can still be improved by adding the consistent attentive layers. The module works because we use the shared weight to extract multiple embeddings' consensus information and fuse them as new embedding, and finally the fusion of multiple meaningful features is helpful for constructing a more complete representation.}

\begin{figure*}[t]
	\begin{center}
		\includegraphics[width=0.85\textwidth]{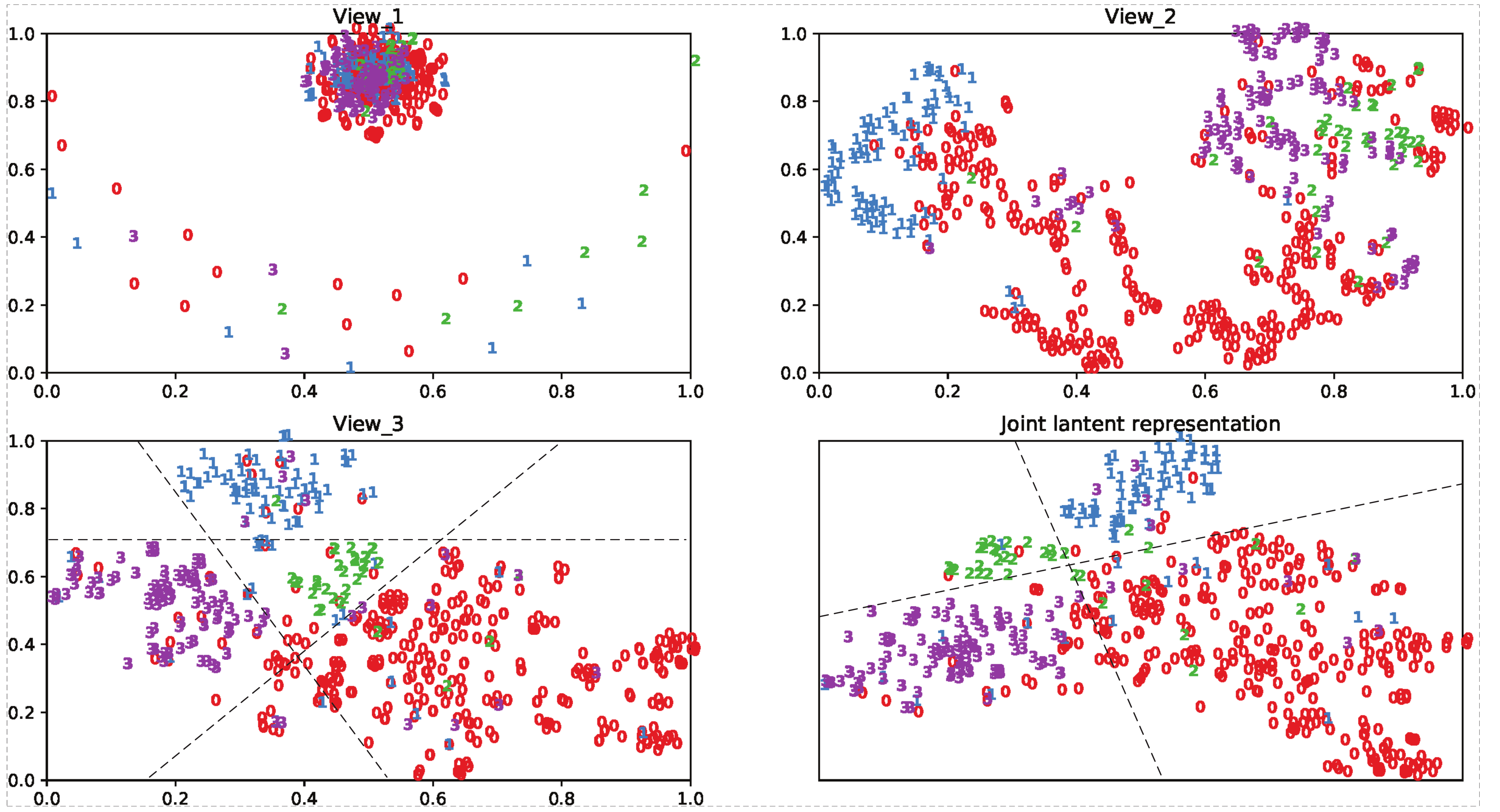}
		\caption{Visualization of original views and fused representation on Prokaryotic: 4 panels represent 2-dim representations of 3 original views and obtained joint latent representation respectively. The data set have 4 clusters, and we label them as 0 - 3, we can find that view 1's original representation is the most chaotic, and view 3's performs best, and it needs 3 decision boundaries to distinct 4 clusters. And for our computed joint latent representation, we only need 2 decision boundaries that can distinct different targets well.} \label{figure_2}
	\end{center}
\end{figure*}

\begin{figure*}[h!]
	\begin{center}
		\includegraphics[width=0.85\textwidth]{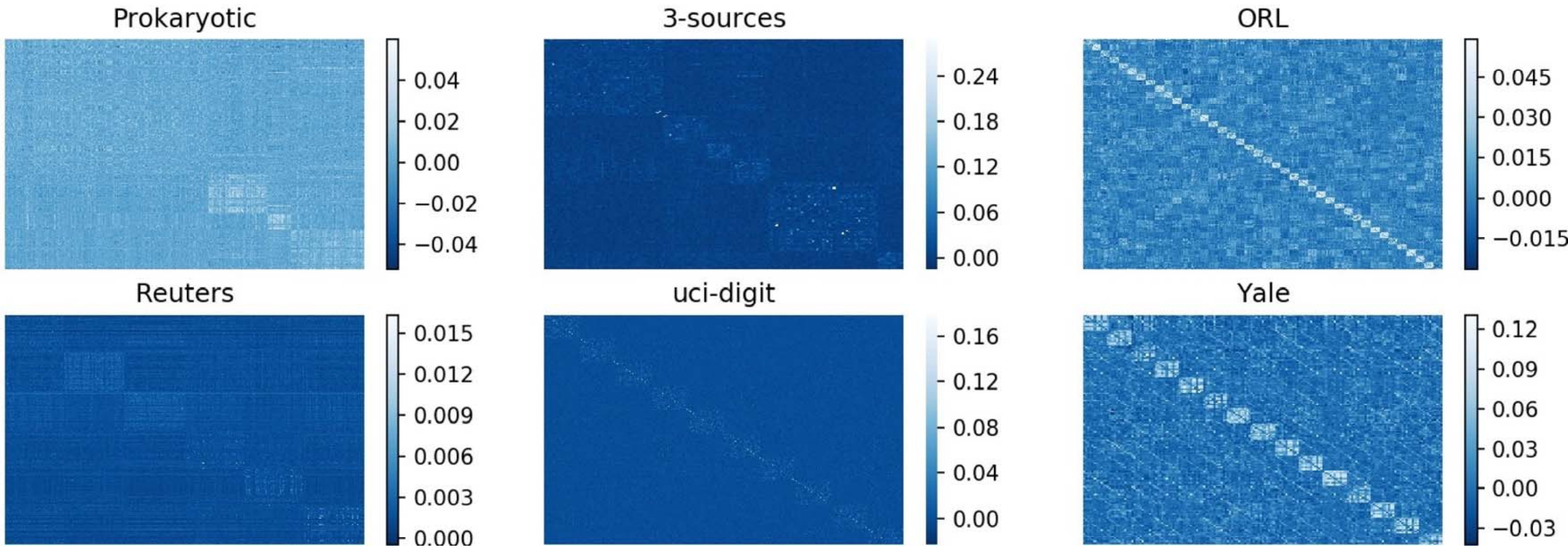}
		\caption{Correlation Diagram of Self-Representation Matrix: we illustrate the correlation heat maps on seven data sets, where lighter color indicates higher relevance. The block numbers represent the number of clusters, and the diagram indicates that each sample has higher relationship with those samples belong to same cluster.} \label{figure_3}
	\end{center}
\end{figure*}

\subsection{Advanced Analysis}
To show the difference between representation computed by AMVDSN and each view's original representation, we reduce $\bf{Z}$ and $\left\{ {{{\bf{X}}^v}} \right\}_{v = 1}^V$ of Prokaryotic to 2-dimensional data by t-sne and visualizes them in Figure \ref{figure_2}. We find that the joint representation has a better decision boundary, especially for cluster 2. Note that for original views, view 3 is the best representation, and if we want to divide them into 4 clusters, at least 3 decision boundaries are needed. And for our obtained joint latent representation, we only need two decision boundaries, and the distinction between clusters is clearer.

Furthermore, according to the theory proposed by \cite{8259470}, the self-representation matrix needs to block diagonal so that it can have good clustering performance. Therefore, we visualize the self-representation $\bf{C}$ of all data sets in Figure \ref{figure_3}, which shows the matrix is block diagonal according to categories. In Figure \ref{figure_3}, lighter colors represent stronger relevance, which means on all data sets we selected, each sample is most closely related to the samples that they belong to the same clusters, and this is the reason that self-representation subspace clustering works well.

\section{Conclusion and Future Works}
\label{sec 5}
We proposed a novel multi-view subspace learning framework based on auto-encoder in this paper. By introducing self-attention mechanism, we can dynamically update views' contribution during training process, and it also makes the fusion process more reasonable. With the joint latent representation computed by neural network, we can achieve state-of-the-art clustering results only using least squares regression subspace learning, which indicates a better feature space will derive better clustering performance in subspace learning. Besides, by introducing shortcut connection, we solve the problem of model degradation. Finally, extensive experiments on real-world data sets demonstrate the effectiveness of our proposed AMVDSN.
 
In the future, more advanced regularization strategies on self-representation matrix $\bf{C}$ like low rank and sparse constraints will be introduced to give more reasonable meanings for deep subspace clustering. \textcolor{black}{On the other hand, refer to the success of the transformer, we can attempt to treat the data of multiple views as a sequence and introduce self-representation learning into the structure of encoder-decoder, which will enhance each view's expression ability by the help of other views and eliminate the corruptions of some disturbed views.} Finally, we find that it's hard to constrain the diagonal entries of $\bf{C}$ equal to zero, we will focus on novel optimal method to handle the problem instead of using some tricks. In general, multi-view subspace clustering is a promising research direction of machine learning, and we will focus on this direction and continue to work for it.

\end{document}